\documentclass[10pt,twocolumn,letterpaper]{article}

\usepackage{cvpr}
\usepackage{times}
\usepackage{epsfig}
\usepackage{graphicx}
\usepackage{amsmath}
\usepackage{amssymb}
\usepackage{epsfig}
\usepackage{graphicx}
\usepackage{amsmath}
\usepackage{amssymb}
\usepackage{algorithm}
\usepackage{algpseudocode}
\usepackage{multirow}
\usepackage{color}
\usepackage{color, colortbl}
\usepackage{authblk}
\definecolor{Gray}{gray}{0.8}
\definecolor{Red}{rgb}{0.95,0.6,0.6}

\usepackage[breaklinks=true,bookmarks=false]{hyperref}

\cvprfinalcopy 

\def\cvprPaperID{****} 

\renewcommand{\paragraph}{\textbf}
\expandafter\def\expandafter\normalsize\expandafter{%
\normalsize\setlength\abovedisplayskip{3pt}}
\expandafter\def\expandafter\normalsize\expandafter{%
\normalsize\setlength\belowdisplayskip{3pt}}

\begin{document}

\title{Deep Sketch Hashing: Fast Free-hand Sketch-Based Image Retrieval}

\author[1]{Li Liu}
\author[2]{Fumin Shen}
\author[1]{Yuming Shen}
\author[3]{Xianglong Liu}
\author[1]{Ling Shao}

\affil[1]{School of Computing Science, University of East Anglia, UK}
\affil[2]{Big Media Computing Center, University of Electronic Science and Technology of China, China}
\affil[3]{School of Computer Science and Engineering, Beihang University, China}
\affil[ ]{\small{\texttt{\{li.liu, yuming.shen, ling.shao\}@uea.ac.uk, fumin.shen@gmail.com, xlliu@nlsde.buaa.edu.cn}}}

\maketitle

\begin{abstract}
Free-hand sketch-based image retrieval (SBIR) is a specific cross-view retrieval task, in which queries are abstract and ambiguous sketches while the retrieval database is formed with natural images. Work in this area mainly focuses on extracting representative and shared   features for  sketches and  natural  images. However, these can neither cope well  with the geometric distortion  between sketches and images nor be feasible for  large-scale SBIR  due to the heavy continuous-valued distance computation.  In this paper,  we speed up SBIR by introducing a novel binary coding method, named \textbf{Deep Sketch Hashing} (DSH), where a semi-heterogeneous deep architecture is proposed and  incorporated into an end-to-end binary coding framework. Specifically, three convolutional neural networks are utilized to encode free-hand sketches, natural images and,  especially, the auxiliary sketch-tokens  which are adopted as bridges to mitigate the sketch-image geometric distortion.
The learned DSH codes can effectively capture the cross-view similarities  as well as the intrinsic semantic correlations between different categories.
To the best of our knowledge, DSH is the first hashing work specifically designed for category-level SBIR with an end-to-end deep architecture. The proposed DSH is comprehensively evaluated on two large-scale datasets of TU-Berlin Extension and Sketchy, and the experiments consistently
show DSH's superior SBIR accuracies over several state-of-the-art methods, while achieving significantly reduced retrieval time and  memory footprint.
\end{abstract}

\vspace{-1ex}
\section{Introduction}
\begin{figure}
  \centering
  \includegraphics[width=0.49\textwidth,height=0.18\textheight]{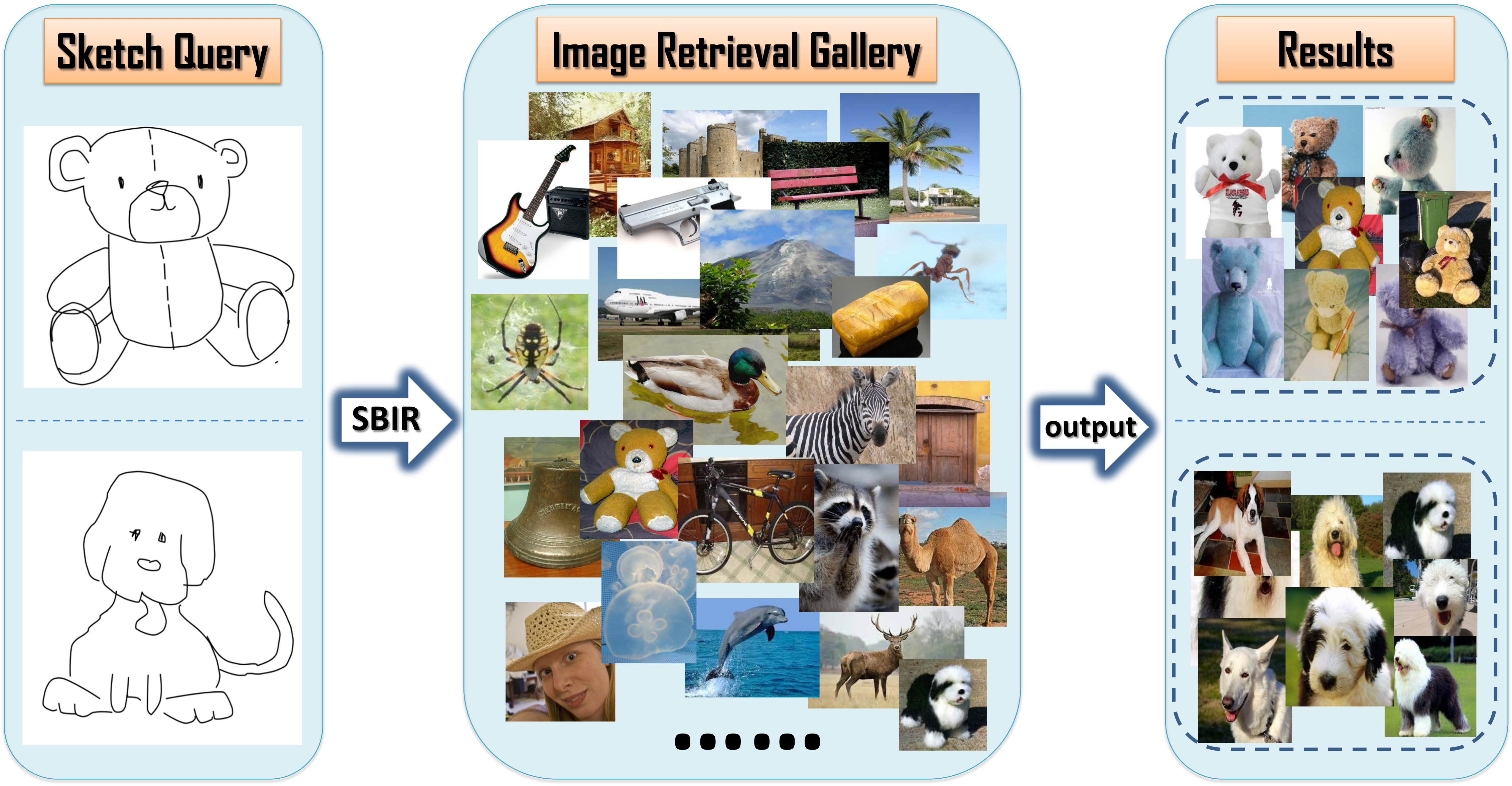}\\
  \caption{An illustration of the SBIR concept in this paper. Given a free-hand query, we aim to retrieve relevant natural images in the same category as the query from the gallery. }\label{intro}
\vspace{-1ex}
\end{figure}

Content-based image retrieval (CBIR) or text-based retrieval (TBR) has played a major role in practical computer vision applications.  In some scenarios, however, if  example queries are not available or it is  difficult to describe them with keywords, what should we do? To address such a problem, sketch-based image retrieval (SBIR) \cite{eitz2011sketch,hu2013performance,saavedra2014sketch,zhou2012sketch,parui2014similarity,saavedrasketch,cao2013sym,eitz2010evaluation,li2016fine,james2014reenact,wang2015sketch,cao2011edgel,cao2010mindfinder,hu2011bag,yusketch,qi2016sketch,sangkloy2016sketchy} has been recently developed and is becoming popular in information retrieval area (as shown in Fig.~\ref{intro}). Compared to traditional retrieval approaches, using a sketch query can more efficiently and precisely express the shape, pose and fine-grained details of the search target, which is intuitive to humans and far more convenient than describing  it with a ``hundred" words in text.

However, SBIR is challenging since humans draw free-hand sketches without any reference but only focus on the salient object structures. As such, the shapes and scales in sketches are usually distorted compared to natural images. To deal with this problem, some studies have attempted to bridge the domain gap between sketches and natural images for SBIR. These methods can be roughly divided into two groups: hand-crafted methods and cross-domain deep learning-based methods.

Hand-crafted SBIR first  generates approximate sketches by extracting  edge or contour maps from the natural images. After that, hand-crafted features (\emph{e.g.}, SIFT \cite{lowe1999object}, HOG \cite{dalal2005histograms}, gradient field HOG \cite{hu2010gradient,hu2013performance}, histogram of edge local orientations (HELO) \cite{saavedra2010improved,saavedra2014sketch} and Learned KeyShapes (LKS) \cite{saavedrasketch}) are extracted for both sketches and edgemaps of natural images, which are then fed into ``Bag-of-Words" (BoW) methods to generate the representations for SBIR. The major limitation of hand-crafted methods is that the domain gap between sketches and natural images cannot be well remedied, as it is difficult to match edge maps to non-aligned sketches with large variations and ambiguity.

To further improve the above domain shift issue, convolutional neural networks (CNNs) \cite{krizhevsky2012imagenet} have  been recently used to  learn domain-transformable features from sketches and images with  end-to-end frameworks \cite{sangkloy2016sketchy,qi2016sketch,yusketch}.
Being able to better handle the domain gap, deep methods typically  achieve higher performance than  hand-crafted ones for both category-level \cite{eitz2011sketch,hu2013performance,saavedra2014sketch,zhou2012sketch,parui2014similarity,saavedrasketch,eitz2010evaluation} and fine-grained \cite{sangkloy2016sketchy,yusketch,li2016fine} SBIR tasks.

Although achieving progress,  current deep SBIR methods are still facing severe challenges. In particular, these methods tend to perform well in the situation that each of the gallery images contains only a single object with a simple contour shape on a clean background  (\eg,  ``Moon", ``Eiffel-tower" and ``Pyramid" in the shape-based Flickr15K dataset \cite{hu2013performance}). In practice, however, objects in gallery images may appear from various viewpoints with relatively complex backgrounds (\emph{e.g.}, a rhinoceros in bushes). In such a case, current methods fail to handle the significant geometric distortions between free-hand sketches and natural images, and result in unsatisfactory performance.


Moreover, less study has been devoted to the searching efficiency of SBIR. Most SBIR techniques are based on applying  nearest neighbor (NN) searches with computational complexity $\mathcal{O}(Nd)$ on  {continuous-valued} features (hand-crafted or deeply learned). Such methods become inappropriate for large-scale SBIR tasks in certain realistic scenarios (\emph{e.g.}, on wearable or mobile devices).  
Therefore, being able to conduct a \emph{{fast}} SBIR on a substantial number of images with \emph{{limited computational and memory resources}} is  crucial for practical applications.

To address the above issues, in this paper, we introduce a novel  \emph{Deep Sketch Hashing (DSH)} framework for the fast free-hand SBIR, which incorporates the learning of binary codes and deep hash functions into a unified framework.
Specifically, DSH speeds up SBIR by embedding sketches and natural images into two sets of compact binary codes, aiming at not only preserving their pairwise semantic similarities, but also leveraging the intrinsic category correlations.  Unlike previous methods with Siamese  \cite{qi2016sketch,wang2015sketch} or triplet CNNs \cite{sangkloy2016sketchy,yusketch} only utilizing images and sketches, we propose a novel semi-heterogeneous deep  architecture including three CNNs, where a unique middle-level network fed with ``sketch-tokens" is developed to effectively diminish the aforementioned geometric distortion between free-hand sketches and natural images.
The contributions of this work mainly include:\vspace{-1ex}
\begin{itemize}
\itemsep0em
\item To the best of our knowledge,  DSH is the first hashing work specifically designed for category-level SBIR, where both binary codes and deep hash functions are learned in a joint end-to-end framework. DSH aims to generate binary codes which can successfully capture  the cross-view relationship (between images and sketches) as well as the intrinsic semantic correlations between different categories.  To this end, an efficient alternating optimization scheme is applied to produce the high-quality hash codes.
\item
A novel semi-heterogeneous deep architecture is developed in DSH as the hash function, where natural images, free-hand sketches and the auxiliary sketch-tokens are fed into  three CNNs (as shown in Fig.~\ref{overall}).  Particularly, natural images and their corresponding sketch-tokens are fed into a heterogeneous late-fusion net, while the CNNs for sketches and sketch-tokens share the same weights during training. As such, the architecture in DSH can better remedy the domain gap between images and sketches compared to previous SBIR deep nets.
\item
The experiments consistently illustrate superior performance of DSH  compared to the state-of-the-art methods, while achieving \emph{significant reduction on both retrieval time and memory load}. 
\end{itemize}

\paragraph{Related Work}
Hashing techniques \cite{gionis1999similarity,liu2015multiview,weiss2008spectral,liu2011hashing,liu2015projection,gong2013iterative,zhu2016deep,liu2016latent,erin2015deep,zhang2016efficient,liu2014discrete,raginsky2009locality,liu2012supervised,shen2015supervised,kulis2009kernelized,liu2016sequential}   have recently been successfully applied to encode high-dimensional features into compact  similarity-preserving  \emph{binary codes}, which enables  extremely fast similarity search by the use of Hamming distances. Inspired by this, some recent SBIR works \cite{bozas2012large,furuya2014hashing,matsui2014sketch2manga,siddiquie2014multi,sun2013indexing,tseng2012sketch} have incorporated  existing hashing methods for efficient retrieval. For instance, LSH \cite{gionis1999similarity} and ITQ \cite{gong2013iterative} are  adopted to  sketch-based image  \cite{bozas2012large}   and 3D model \cite{furuya2014hashing} retrieval tasks, respectively. In fact, among various hashing methods, cross-modality hashing  \cite{lin2015semantics,zhang2014large,zhou2014latent,kumar2011learning,liu2017sequential,bronstein2010data,DBLP:conf/sigmod/SongYYHS13,zhu2013linear,zhen2012co,ding2014collective,jiang2016deep,caodeep,cao2016correlation}, which learns binary codes by preserving the correlations between heterogeneous representations from different modalities, are  more related to SBIR problems. However, all of the above hashing techniques are not specifically designed for SBIR and neglect the intrinsic relationship between free-hand sketches and natural images, resulting in unsatisfactory performance.

In the next section, we will introduce the detailed architecture of our deep hash nets in DSH, then elaborate on our hashing objective function.

\section{Deep Sketch Hashing}
To help better understand this section, we first introduce some notation.
Let $\mathcal{O}_1=\{\mathbf{I}_i,\mathbf{Z}_i\}^{n_1}_{i=1}$, where $\mathbf{I}_i$ is a natural image and $\mathbf{Z}_i$ is its corresponding sketch-token computed from $\mathbf{I}_i$;  $\mathcal{O}_2=\{\mathbf{S}_j\}^{n_2}_{j=1}$ be the set of free-hand sketches $\mathbf{S}_j$; and  $n_1$ and $n_2$ indicate the numbers of the samples in $\mathcal{O}_1$ and $\mathcal{O}_2$, respectively. Additionally, define the label matrix $\mathbf{Y}^{I}=\{\mathbf{y}^{I}_{i}\}^{n_1}_{i=1}\in \mathbb{R}^{C\times n_1}$, where $y_{ci}^{I}=1$ if $\{\mathbf{I}_i,\mathbf{Z}_i\}$ belongs to class $c$ and $0$ otherwise; $\mathbf{Y}^{S}=\{\mathbf{y}^{S}_{j}\}_{j=1}^{n_2}\in \mathbb{R}^{C\times n_2}$ for sketches is defined in the same way.  We aim to learn two sets of $m$-bit binary codes $\mathbf{\mathbf{B}}^{I}=\{\mathbf{b}^{I}_{i}\}^{n_1}_{i=1}\in\{-1,1\}^{m\times n_1}$   and $\mathbf{B}^{S}=\{\mathbf{b}^{S}_{j}\}^{n_2}_{j=1}\in\{-1,1\}^{m\times n_2}$ for $\mathcal{O}_1$ and $\mathcal{O}_2$, respectively.
\subsection{Semi-heterogeneous Deep Architecture}\label{21}
\begin{figure}
  \centering
  \includegraphics[width=0.495\textwidth]{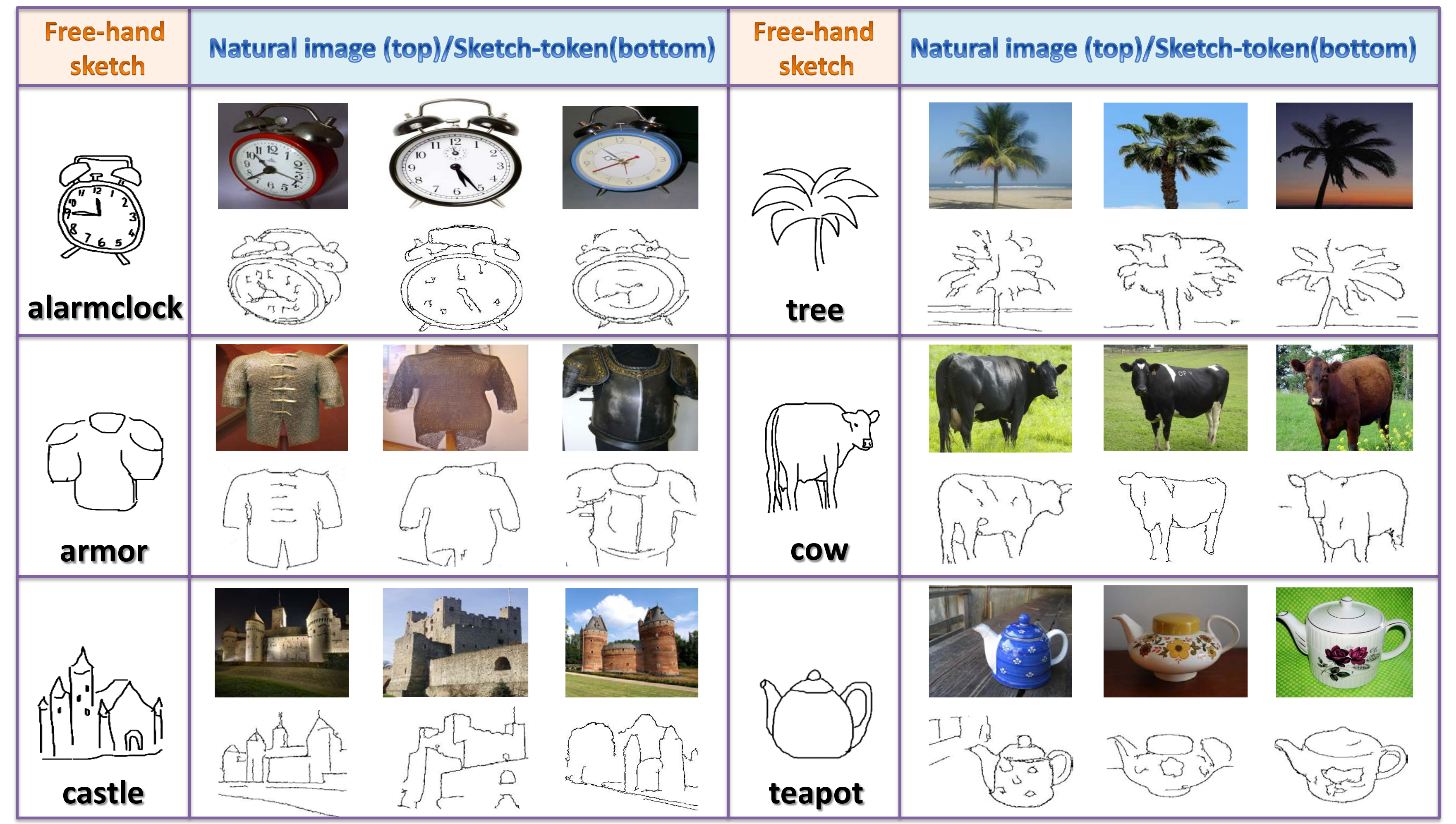}\\
  \caption{Illustration of our DSH inputs: free-hand sketches, natural images and corresponding sketch-tokens.  Sketch-tokens have similar stroke patterns and appearance to free-hand sketches.
  }\label{example}
\vspace{-0.5ex}
\end{figure}

As previously stated, SBIR is a very challenging task due to large geometric distortion between sketches and images.
Inspired by \cite{lim2013sketch,saavedrasketch}, in this work, we propose to adopt an auxiliary image representation  as a  bridge to mitigate the geometric distortion between sketch and natural images.
In particular,  a set of edge structures are detected from natural images,   called ``sketch-tokens'',  using supervised middle-level information in the form of \emph{hand-drawn sketches}.  In practice, given an image we will get an initial sketch-token, where each  pixel is assigned a score for the likeliness of it being a contour point. We then  use 60\% of the maximum score (same as \cite{saavedrasketch}) to threshold each pixel and obtain the final sketch-tokens as shown in Fig.~\ref{example}.

\vspace{1ex}
Sketch-tokens have two advantages: (1) they reflect only essential edges of natural images without detailed texture information; (2) unlike ordinary edgemaps (\emph{e.g.}, Canny), they have very similar stroke patterns and appearance to free-hand sketches. Next, we will show how to design the DSH architecture with the help of sketch-tokens.

\begin{figure*}
  \centering
  \includegraphics[width=1\textwidth,height=0.3\textheight]{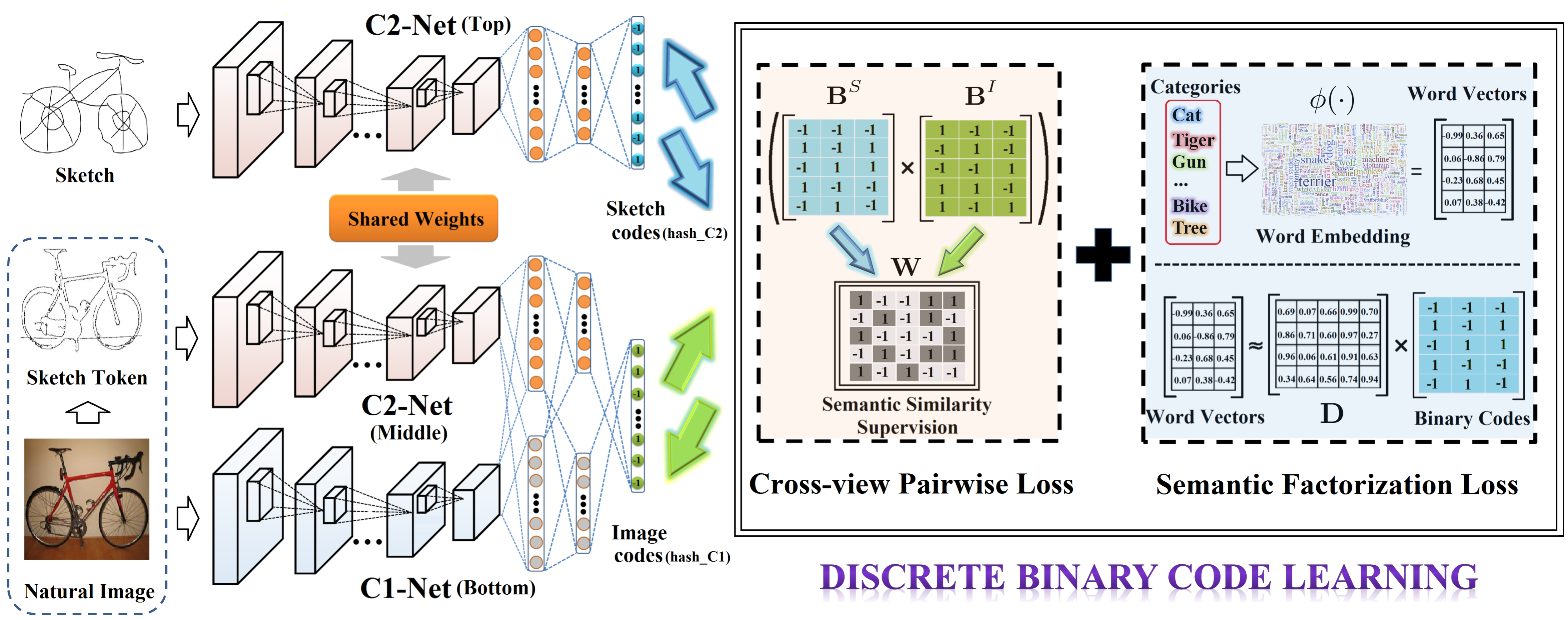}\\
  \caption{The illustration of the main idea of the proposed DSH. Specifically, we integrate a convolutional neural network and discrete binary code learning  into a unified end-to-end framework which can be effectively optimized in an alternating manner.}\label{overall}
\vspace{-2.5ex}
\end{figure*}

\vspace{1ex}
We propose  a novel semi-heterogeneous deep architecture, where three CNNs are developed as hash functions to encode free-hand sketches, natural images and auxiliary  sketch-tokens into binary codes. As shown in Fig.~\ref{overall}, the  DSH framework includes the following two  parts:

\vspace{2ex}
\textbf{1) Cross-weight Late-fusion Net:} A heterogeneous net with two parallel CNNs is developed, termed C1-Net (Bottom) and C2-Net (Middle). Particularly, C1-Net (bottom) is slightly modified from AlexNet \cite{krizhevsky2012imagenet} containing 5 convolutional (\emph{conv}) layers and 2 fully connected (\emph{fc}) layers for natural image inputs, while C2-Net is configured with 4 convolutional layers and 2 fully connected layers for corresponding sketch-token inputs. The detailed parameters are listed in Table~\ref{table:t1}.  Inspired by the recent multimodal deep framework \cite{rastegar2016mdl}, we  connected the \emph{pooling3}, \emph{fc\_a}, \emph{fc\_b} of both C1-Net (Bottom) and C2-Net (Middle) with cross-weights. In this way, we  exploit high-level interactions between two nets to maximize the mutual information across both modalities, while the information from each individual net is also preserved. Finally, similar to \cite{lin2015semantics,ding2014collective},
we late-fuse the C1-Net (Bottom) and C2-Net (Middle) into a unified binary coding layer $\texttt{\textbf{hash\_C1}}$ so that the learned codes can fully benefit from both natural images and their corresponding sketch-tokens.

\begin{table}[t]
\newcommand{\tabincell}[2]{\begin{tabular}{@{}#1@{}}#2\end{tabular}}
\caption{The detailed configuration of the proposed DSH. }
\begin{center}
\resizebox{0.45\textwidth}{!}{
\begin{tabular}{|c|c|c|c|c|c|}
\hline
\textbf{Net} & \textbf{Layer} & \textbf{Kernel Size} & \textbf{Stride} & \textbf{Pad}& \textbf{Output} \\
\hline
\hline
\multirow{14}{5em}{\tabincell{c}{\textbf{C1-Net}\\(\textbf{Natural}\\\textbf{Image})}}
&input &- &- &- &3$\times$227$\times$227 \\\cline{2-6}
&\emph{conv1} &11$\times$11 &4 &0 &96$\times$55$\times$55 \\
&\emph{pooling1} &3$\times$3 &2 &0 &96$\times$27$\times$27 \\\cline{2-6}
&\emph{conv2} &5$\times$5 &1 &2 &256$\times$27$\times$27 \\
&\emph{pooling2} &3$\times$3&2 &0 &256$\times$13$\times$13  \\\cline{2-6}
&\emph{conv3} &3$\times$3 &1 &1 &384$\times$13$\times$13 \\
&\emph{conv4} &3$\times$3 &1 &1 &384$\times$13$\times$13 \\
&\emph{conv5} &3$\times$3 &1 &1 &384$\times$13$\times$13 \\
&\emph{pooling3} &3$\times$3 &2 &1 &256$\times$7$\times$7 \\\cline{2-6}
&\emph{fc\_a} &7$\times$7 &1 &0 &4096$\times$1$\times$1  \\
&\emph{fc\_b} &1$\times$1 &1 &0 &1024$\times$1$\times$1  \\\cline{2-6}
&$\texttt{\textbf{hash\_C1}}$ &1$\times$1 &1 &0 &$m$ $\times$1$\times$1 \\\cline{2-6}
\hline
\hline
\multirow{12}{5em}{\tabincell{c}{\textbf{C2-Net}\\(\textbf{Free-hand}\\ \textbf{sketch/}\\\textbf{Sketch-}\\\textbf{tokens} )}}
&input &- &- &- &1$\times$200$\times$200 \\\cline{2-6}
&\emph{conv1} &14$\times$14 &3 &0 &64$\times$63$\times$63 \\
&\emph{pooling1} &3$\times$3 &2 &0 &64$\times$31$\times$31 \\\cline{2-6}
&\emph{conv2\_1} &3$\times$3 &1 &1 &128$\times$31$\times$31 \\
&\emph{conv2\_2} &3$\times$3 &1 &1 &128$\times$31$\times$31 \\
&\emph{pooling2} &3$\times$3&2 &0 &128$\times$15$\times$15  \\\cline{2-6}
&\emph{conv3\_1} &3$\times$3 &1 &1 &256$\times$15$\times$15 \\
&\emph{conv3\_2} &3$\times$3 &1 &1 &256$\times$15$\times$15 \\
&\emph{pooling3} &3$\times$3 &2 &0 &256$\times$7$\times$7 \\\cline{2-6}
&\emph{fc\_a} &7$\times$7 &1 &0 &4096$\times$1$\times$1  \\
&\emph{fc\_b} &1$\times$1 &1 &0 &1024$\times$1$\times$1  \\\cline{2-6}
&$\texttt{\textbf{hash\_C2}}$ &1$\times$1 &1 &0 &$m$ $\times$1$\times$1 \\\cline{2-6}
\hline
\end{tabular}
}
\end{center}
\label{table:t1}
\vspace{-3ex}
\end{table}
\vspace{0.3em}
\textbf{2) Shared-weight Sketch Net:} For free-hand sketch inputs, we develop the C2-Net (Top) with configurations shown in Table~\ref{table:t1}.
Specifically, considering the similar characteristics and implicit correlations existing between sketch-tokens and free-hand sketches as mentioned above, we design a Siamese architecture for C2-Net (Middle) and C2-Net (Top) to share the same deep weights in \emph{conv}  and \emph{fc} layers during the optimization (see in Fig.~\ref{overall}). As such, the hash codes of free-hand sketches learned via the shared-weight net (from $\texttt{\textbf{hash\_C2}}$)  will mitigate the geometric difference between  images and sketches during SBIR.

\vspace{0.3em}
 \noindent\textbf{Deep Hash Functions:} Denote by $\Theta_{1}$ the deep weights in C1-Net (Bottom) and  $\Theta_{2}$ the shared weights in C2-Net (Middle) and C2-Net (Top). For  natural images and their sketch-tokens, we form  the deep hash function  $\mathbf{B}^{I} = \operatorname{sign}(\mathbf{F}_1(\mathcal{O}_1;\Theta_{1},\Theta_{2}))$  from the cross-weight late-fusion net of C1-Net (Bottom) and C2-Net (Middle). Similarly, the shared-weight sketch net (\emph{i.e.}, C2-Net (Top)) is regarded as the hash function $\mathbf{B}^{S}=\operatorname{sign}(\mathbf{F}_2(\mathcal{O}_2;\Theta_{2}))$ for free-hand sketches. In this way, hash codes learned from the above deep hash functions can lead to more reasonable SBIR, especially when a significant sketch-image distortion exists. Next, we will introduce the DSH objective of joint learning of binary codes and hash functions.

\subsection{Objective Formulation of DSH}\label{22}


\textbf{1) Cross-view Pairwise Loss:}
We first define the cross-view similarity matrix of $\mathcal{O}_1$ and $\mathcal{O}_2$ as $\mathbf{W}\in\mathbb{R}^{n_1\times n_2}$, where the element of $\mathbf{W}_{ij}$ denotes the cross-view similarity between $\{\mathbf{\mathbf{I}}_i,\mathbf{\mathbf{Z}}_i\}$ and $\mathbf{S}_j$.  The inner product of learned $\mathbf{B}^{I}$ and $\mathbf{B}^{S}$ should sufficiently approximate the similarity matrix $\mathbf{W}$. Thus, we consider the following problem:\vspace{-0.5ex}
\begin{small}
\begin{align}\label{e1}
&\underset{\mathbf{B}^{I},\mathbf{B}^{S}}{\min}\mathcal{J}_1:=||\mathbf{W}\odot m-{\mathbf{B}^{I}}^\top\mathbf{B}^{S}||^{2},\\ \nonumber
&\mathrm{s.t.}\ \mathbf{B}^{I}\in\{-1,+1\}^{m\times n_1},\ \mathbf{B}^{S}\in\{-1,+1\}^{m\times n_2},
\end{align}
\end{small}%
where $||\cdot||$ is the Frobenius norm and $\odot$ is the element-wise product. The cross-view similarity matrix $\mathbf{W}$ can be defined by semantic label information as $\mathbf{W}_{ij}=1$ if $\mathbf{y}^I_i=\mathbf{y}^S_j$ and $-1$ otherwise.
By Eq.(\ref{e1}), the binary codes of natural images and  sketches from the same category will be pulled  as close as possible and  pushed far away otherwise.

\vspace{1ex}
\textbf{2) Semantic Factorization Loss:}
Beyond the cross-view similarity, we also consider preserving the intra-set semantic relationships for both the image set $\mathcal{O}_1$ and the sketch set $\mathcal{O}_2$. However, the given 0/1 label matrices $\mathbf{Y}^{I}$ and $\mathbf{Y}^{S}$  can only provide binary measurements (\ie, the samples belong to the same category or not), which causes all different categories to have equivalent distance (\eg, ``\emph{cheetah}" will be as different from  ``\emph{tiger}" as from ``\emph{dolphin}").  Thus, directly using such discrete label information will implicitly make all categories independent and discards the latent correlation of high-level semantics. \vspace{2mm}

Inspired by the recent development of word embeddings \cite{mikolov2013distributed}, in this paper,  we overcome the above drawback by utilizing the NLP word-vector toolbox\footnote{https://code.google.com/archive/p/word2vec/. The model is trained from the first billion characters from Wikipedia.} to map the independent labels into the high-level semantic space. As such, the intrinsic semantic correlation among different labels can be quantitatively measured and captured (\emph{e.g.}, the semantic embedding of ``\emph{cheetah}"  will be  closer to ``\emph{tiger}"  but further from ``\emph{dolphin}"). As semantic embeddings  intentionally guide the learning of high-quality binary codes,  we optimize the following semantic factorization problem
\begin{small}
\begin{align}\label{e3}
&\underset{\mathbf{B}^{I},\mathbf{B}^{S}}{\min}\mathcal{J}_2:=||\phi(\mathbf{Y}^{I})-\mathbf{D}\mathbf{B}^{I}||^{2}+||\phi(\mathbf{Y}^{S})-\mathbf{D}\mathbf{B}^{S}||^{2},\\ \nonumber
&\mathrm{s.t.}\ \mathbf{B}^{I}\in\{-1,+1\}^{m\times n_1},\ \mathbf{B}^{S}\in\{-1,+1\}^{m\times n_2},
\end{align}
\end{small}%
where $\phi(\cdot)$ is the word embedding model, $\phi(\mathbf{Y}^{I})\in\mathbb{R}^{d\times n_1}$ and $\phi(\mathbf{Y}^{S})\in\mathbb{R}^{d\times n_2}$, $d=1000$ is the dimension of word embedding. $\mathbf{D}\in\mathbb{R}^{d\times m}$ is the shared basis of the semantic factorization for both views. Note that the shared basis we used helps to preserve the latent semantic correlations which also benefits cross-view code learning in SBIR.

\vspace{2ex}
\textbf{Final Objective Function:} Unlike previous hashing methods using continuous-relaxation during code learning, we keep the binary constraints in the DSH optimization.  By recalling Eq.(\ref{e1}) and  Eq.(\ref{e3}), we obtain our final objective function:
\begin{small}
\begin{align}\label{e4}
&\underset{\mathbf{B}^{I},\mathbf{B}^{S},\mathbf{D}^{I},\mathbf{D}^{S},\Theta_{1},\Theta_{2}}{\min}\mathcal{J}:=||\mathbf{W}\odot m-{\mathbf{B}^{I}}^\top\mathbf{B}^{S}||^{2}\\  \nonumber
&+\lambda(||\phi(\mathbf{Y}^{I})-\mathbf{D}\mathbf{B}^{I}||^{2}+||\phi(\mathbf{Y}^{S})-\mathbf{D}\mathbf{B}^{S}||^{2})\\\nonumber
&+\gamma(||\mathbf{F}_{1}(\mathcal{O}_1;\Theta_{1},\Theta_{2})-\mathbf{B}^{I}||^{2}+||\mathbf{F}_{2}(\mathcal{O}_2;\Theta_{2})-\mathbf{B}^{S}||^{2}),\\\nonumber
&\mathrm{s.t.}\ \mathbf{B}^{I}\in\{-1,+1\}^{m\times n_1},\ \mathbf{B}^{S}\in\{-1,+1\}^{m\times n_2} \nonumber.
\end{align}
\end{small}%
Here, $\lambda>0$ and $\gamma>0$ are the balance parameters.
The last two regularization terms aim to minimize  the quantization loss between binary codes $\mathbf{B}^{I}$, $\mathbf{B}^{S}$ and deep hash functions $\mathbf{F}_1(\mathcal{O}_1;\Theta_{1},\Theta_{2})$, $\mathbf{F}_2(\mathcal{O}_2;\Theta_{2})$.
Similar regularization terms are also used in \cite{shen2015learning,liu2014discrete} for effective hash code learning. Next, we will  elaborate on how to optimize problem~\eqref{e4}.


\section{Optimization}
It is clear that problem~\eqref{e4} is non-convex and non-smooth, which is in general an NP-hard problem due to the binary constraints. To address this, we propose an alternating optimization based algorithm, which sequentially updates $\mathbf{D}$, $\mathbf{B^{I}}$, $\mathbf{B^{S}}$ and deep hash functions $\mathbf{F}_1/\mathbf{F}_2$ in an iterative fashion. In practice, we first pre-train C1-Net (Bottom) and C2-Net (Top) as classification nets using natural images and sketches with corresponding semantic labels. After that, pre-trained models will be applied in our semi-heterogeneous deep model as in Fig.~\ref{overall} and then optimized with the following alternating  steps.
\vspace{-2ex}
\subparagraph{$\mathbf{D}$ Update Step.} By fixing all variables except for $\mathbf{D}$, Eq.(\ref{e4}) shrinks to a classic quadratic regression problem
\begin{small}
\begin{equation}\label{e5}
\underset{\mathbf{D}}{\min} \ ||\phi(\mathbf{Y}^{I})-\mathbf{D}\mathbf{B}^{I}||^{2}+||\phi(\mathbf{Y}^{S})-\mathbf{D}\mathbf{B}^{S}||^{2},
\end{equation}
\end{small} %
which  can be solved  analytically as
\begin{small}
\begin{equation}\label{e6}
\mathbf{D}=(\phi(\mathbf{Y}^{I}){\mathbf{B}^{I}}^{\top}+\phi(\mathbf{Y}^{S}){\mathbf{B}^{S}}^{\top})(\mathbf{B}^{I}{\mathbf{B}^{I}}^{\top}+\mathbf{B}^{S}{\mathbf{B}^{S}}^{\top})^{-1}.
\end{equation}
\end{small}%
\vspace{-5ex}
\subparagraph{$\mathbf{B}^{I}$ Update Step.} By fixing all other variables, we optimize  $\mathbf{B}^{I}$ by the following equation
\begin{small}
\begin{align}\label{e8}
&\underset{\mathbf{B}^{I}}{\min}\ ||\mathbf{W}\odot m-{\mathbf{B}^{I}}^\top\mathbf{B}^{S}||^{2}
+\lambda||\phi(\mathbf{Y}^{I})-\mathbf{D}\mathbf{B}^{I}||^{2}\\
&+\gamma||\mathbf{F}_{1}(\mathcal{O}_1;\Theta_{1},\Theta_{2})-\mathbf{B}^{I}||^{2},\nonumber \\
&\mathrm{s.t.}\ \mathbf{B}^{I}\in\{-1,+1\}^{m\times n_1} \nonumber.
\end{align}
\end{small}%
We further rewrite \eqref{e8} as
\begin{small}
\begin{align}\label{e9}
&\underset{\mathbf{B}^{I}}{\min}\ ||{\mathbf{B}^{I}}^\top\mathbf{B}^{S}||^{2}
+\lambda||{\mathbf{B}^{I}}^{\top}{\mathbf{D}}^{\top}||^{2}-2\operatorname{trace}({\mathbf{B}^{I}}^\top \mathbf{R}),\\
&\mathrm{s.t.}\ \mathbf{B}^{I}\in\{-1,+1\}^{m\times n_1} \nonumber,
\end{align}
\end{small}%
where \begin{small}
$\mathbf{R}=\mathbf{B}^{S}(\mathbf{W}^{\top}\odot m)+\lambda {\mathbf{D}}^{\top}\phi(\mathbf{Y}^{I})+\gamma\mathbf{F}_{1}(\mathcal{O}_1;\Theta_{1},\Theta_{2})$ and
\end{small}
\begin{small}
$||\mathbf{B}^{I}||^{2}=mn_{1}$.
\end{small}%

\begin{algorithm}[t]
\renewcommand{\algorithmicrequire}{\textbf{Input:}}  
\renewcommand{\algorithmicensure}{\textbf{Output:}} 
\small
  \caption{\ \ \ \quad Deep Sketch Hashing (DSH)}
  \label{alg:Framwork}
  \begin{algorithmic}[1]
    \Require
      Set of pairs of natural images  and corresponding sketch-tokens $\mathcal{O}_1=\{\mathbf{I}_i\,\mathbf{Z}_i\}_{i=1}^{n_{1}}$;
      Free-hand sketch set $\mathcal{O}_2=\{\mathbf{S}_J\}_{J=1}^{n_{2}}$;
      The label information $\{\mathbf{y}^{I}_{i}\}_{i=1}^{n_1}$ and $\{\mathbf{y}^{S}_{j}\}_{j=1}^{n_2}$;
      Total epochs $T$ of deep optimization.
    \Ensure
      Deep hash functions $\mathbf{F}_1(\mathcal{O}_1;\Theta_{1},\Theta_{2})$ and $\mathbf{F}_2(\mathcal{O}_2;\Theta_{2})$.
    \State Randomly initialize $\{\mathbf{b}^{I}_{i}\}^{n_1}_{i=1}\in\{-1,+1\}^{m\times n_1}$ and $\{\mathbf{b}^{S}_{j}\}^{n_2}_{j=1}\in\{-1,+1\}^{m\times n2}$ for the entire training set; construct cross-view similarity matrix $\mathbf{W}\in\mathbb{R}^{n_1\times n_2}$.
    \State \textbf{For} $t=1,\ldots,T$ epoch \textbf{do}
    \State ~~~Update $\mathbf{D}$ according to Eq.(\ref{e6});
    \State ~~~Update $\mathbf{B}^{I}$ and $\mathbf{B}^{S}$ according to Eq.(\ref{e11});
    \State ~~~Update the deep parameters \{$\Theta_{1}$,$\Theta_{2}$\} by $t^{th}$ epoch data;
    \State \textbf{End}
  \end{algorithmic}
  \label{aa1}
\end{algorithm}

It is challenging to directly optimize $\mathbf{B}^{I}$ with \emph{discrete constraints}.
Inspired by the \emph{discrete cyclic coordinate descent (DCC)}  \cite{shen2015supervised}, we learn each row of $\mathbf{B}^{I}$ by fixing all other $m-1$ rows, \ie,  each time we only optimize one single bit of all $n_1$ samples. We denote $\mathbf{\widehat{b}}^{I}_{k}$, $\mathbf{\widehat{b}}^{S}_{k}$, $\mathbf{\widehat{r}}_{k}$ and $\mbox{$\mathbf{\widehat{d}}_{k}$}^{\top}$  as the $k^{th}$ rows of $\mathbf{B}^{I}$, $\mathbf{B}^{S}$, $\mathbf{R}$ and $\mathbf{D}^{\top}$ respectively, $k=1,\ldots,m$. For convenience, we also have
\begin{equation}\label{e222}
\small
\vspace{-1ex}
\left\{
\begin{array}{lll}
\mathbf{\widehat{B}}^{I}_{\neg k}&=&[\mbox{$\mathbf{\widehat{b}}^{I}_{1}$}^{\top},\ldots,\mbox{$\mathbf{\widehat{b}}^{I}_{k-1}$}^{\top},\mbox{$\mathbf{\widehat{b}}^{I}_{k+1}$}^{\top},\ldots,\mbox{$\mathbf{\widehat{b}}^{I}_{m}$}^{\top}]^{\top},\\
\mathbf{\widehat{B}}^{S}_{\neg k}&=&[\mbox{$\mathbf{\widehat{b}}^{S}_{1}$}^{\top},\ldots,\mbox{$\mathbf{\widehat{b}}^{S}_{k-1}$}^{\top},\mbox{$\mathbf{\widehat{b}}^{S}_{k+1}$}^{\top},\ldots,\mbox{$\mathbf{\widehat{b}}^{S}_{m}$}^{\top}]^{\top},\\
\mathbf{\widehat{D}}_{\neg k}&=&[\mbox{$\mathbf{\widehat{d}}_{1}$},\ldots,\mbox{$\mathbf{\widehat{d}}_{k-1}$},\mbox{$\mathbf{\widehat{d}}_{k+1}$},\ldots,\mbox{$\mathbf{\widehat{d}}_{m}$}].
\end{array}\right.
\end{equation}%
It is not difficult to show Eq.\eqref{e9} can be rewritten w.r.t. $\mathbf{\widehat{b}}^{I}_{k}$ as
\begin{small}
\begin{align}\label{e10}
&\underset{\widehat{\mathbf{b}}^{I}_{k}}{\min}\ \widehat{\mathbf{b}}^{I}_{k}(\mbox{$\mathbf{\widehat{B}}^{I}_{\neg k}$}^{\top}\widehat{\mathbf{B}}^{S}_{\neg k}\mbox{$\widehat{\mathbf{b}}^{S}_{k}$}^\top+\lambda
\mbox{$\mathbf{\widehat{B}}^{I}_{\neg k}$}^{\top}\mbox{$\mathbf{\widehat{D}}_{\neg k}$}^{\top}\mathbf{\widehat{d}}_{k}-{\mathbf{\widehat{r}}_{k}}^\top),\\
&\mathrm{s.t.}\ \mathbf{\widehat{b}}^{I}\in\{-1,+1\}^{1\times n_1} \nonumber.
\end{align}
\end{small}%
Thus, the closed-form solution for the $k^{th}$ row of $\mathbf{B}^{I}$ can be obtained by
\begin{small}
\begin{equation}\label{e11}
\mathbf{\widehat{b}}^{I}_{k}=\operatorname{sign}({\mathbf{\widehat{r}}_{k}}-\mbox{$\widehat{\mathbf{b}}^{S}_{k}$}\mbox{$\widehat{\mathbf{B}}^{S}_{\neg k}$}^{\top}\mbox{$\mathbf{\widehat{B}}^{I}_{\neg k}$}
-\lambda\mbox{$\mathbf{\widehat{d}}_{k}$}^{\top}\mbox{$\mathbf{\widehat{D}}_{\neg k}$}\mbox{$\mathbf{\widehat{B}}^{I}_{\neg k}$}).
\end{equation}
\end{small}%
In this way, the binary codes $\mathbf{B}^{I}$ can be optimized bit by bit  and finally reach a stationary point.

\vspace{-0.5ex}
\subparagraph{$\mathbf{B}^{S}$ Update Step.} By fixing all other variables, we learn hash code $\mathbf{B}^{S}$ with a similar formulation to Eq.(\ref{e11}).

\vspace{2.5ex}
\paragraph{$\Theta_{1}$ and $\Theta_{2}$ Update Step.} Once $\mathbf{B}^{I}$ and $\mathbf{B}^{S}$ are obtained, we update parameters $\Theta_1$ and $\Theta_2$ of  C1-Net and C2-Net according to the following \emph{Euclidean loss}:\vspace{1ex}
\begin{small}
\begin{equation}\label{e12}
\underset{\Theta_{1},\Theta_{2}}{\min}\ \mathcal{L}:=||\mathbf{F}_{1}(\mathcal{O}_1;\Theta_{1},\Theta_{2})-\mathbf{B}^{I}||^{2}+||\mathbf{F}_{2}(\mathcal{O}_2;\Theta_{2})-\mathbf{B}^{S}||^{2}.
\vspace{1ex}
\end{equation}
\end{small}%
By first computing the partial gradients  $\frac{\partial \mathcal{L}}{\partial \mathbf{F}_1(\Theta_{1},\Theta_{2})}$ and $\frac{\partial \mathcal{L}}{\partial \mathbf{F}_2(\Theta_{2})}$, we can obtian $\frac{\partial \mathcal{L}}{\partial (\Theta_{1},\Theta_{2})}$ by the chain rule. We then use the standard mini-batch back-propagation (BP) scheme to  simultaneously update $\Theta_{1}$ and $\Theta_{2}$ for our entire deep architecture. In practice, the above procedure can be easily achieved by deep learning toolboxes (\eg, Caffe \cite{jia2014caffe}).
 \begin{figure}
  \centering
  \includegraphics[width=0.485\textwidth]{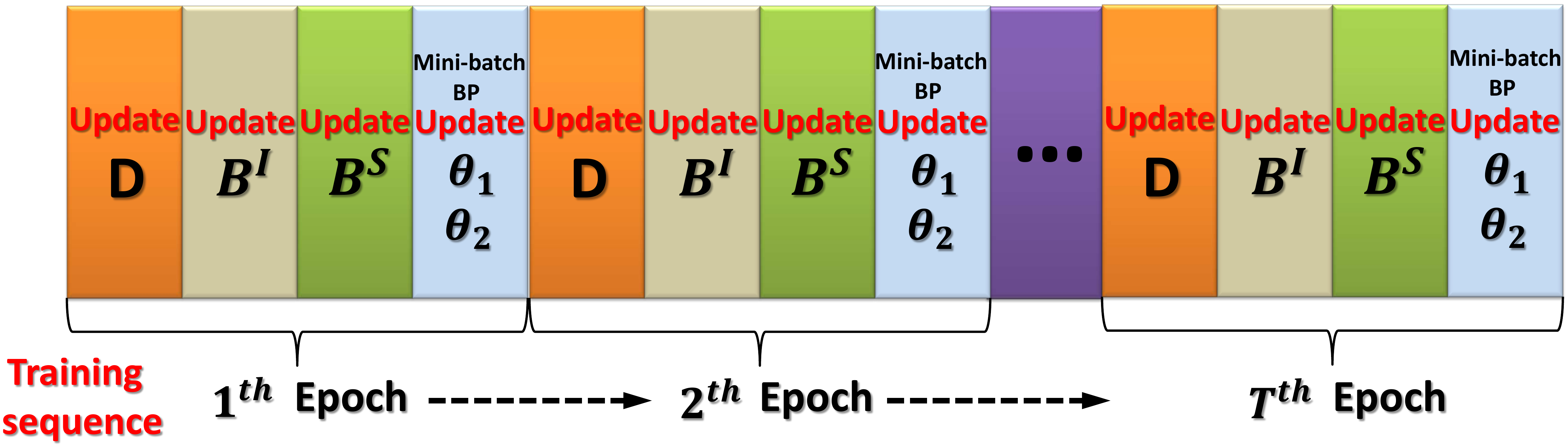}\\
  \caption{The illustration of DSH alternating optimization scheme.}\label{learn}
\end{figure}

\begin{table*}[t]
\center
\newcommand{\tabincell}[2]{\begin{tabular}{@{}#1@{}}#2\end{tabular}}
\caption{Comparison with previous SBIR methods (MAP, Precision@200, Retrieval time ($\mathrm{s}$)/query and Memory load ($\mathrm{MB}$)) on both datasets. Apart from DSH producing binary codes, the continuous-value feature representations are utilized by all other methods.}
\resizebox{0.999\textwidth}{!}{
\begin{tabular}{|c|c||c|c|c|c||c|c|c|c|}
\hline
\multirow{2}{*}{\textbf{Methods}} &\multirow{2}{*}{\textbf{Dimension}}  &\multicolumn{4}{c||}{\textbf{TU-Berlin} \textbf{Extension}}&\multicolumn{4}{c|}{\textbf{Sketchy}} \\
\cline{3-10}
& & \textbf{MAP} & \textbf{\tabincell{c}{Precision\\@200}}&\textbf{\tabincell{c}{Retrieval time\\ per query ($\mathrm{s}$)}} & \tabincell{c}{\textbf{Memory load($\mathrm{MB}$)}\\(204,489 gallery images)}&\textbf{MAP} & \textbf{\tabincell{c}{Precision\\@200}}&\textbf{\tabincell{c}{Retrieval time\\ per query  ($\mathrm{s}$)}}  & \tabincell{c}{\textbf{Memory load($\mathrm{MB}$)}\\(73,002 gallery images)}\\
\hline
HOG \cite{dalal2005histograms} &1296 &0.091 &0.120 &1.43  &$2.02\times10^3$  &0.115 &0.159 & 0.53 &$7.22\times10^2$ \\
GF-HOG \cite{hu2010gradient} &3500 &0.119 &0.148 & 4.13  &$5.46\times10^3$ &0.157 &0.177& 1.41 &$1.95\times10^3$ \\
SHELO \cite{saavedra2014sketch} &1296 & 0.123&0.155 &1.44  & $2.02\times10^3$  &0.161&0.182 &0.50 &$7.22\times10^2$ \\
LKS \cite{saavedrasketch}&1350 &0.157 &0.204 & 1.51 &$2.11\times10^3$   &0.190 &0.230 &0.56 &$7.52\times10^2$ \\
\hline
Siamese CNN \cite{qi2016sketch} &64 &0.322 &0.447 &  7.70$\times 10^{-2}$  &99.8  &0.481 &0.612 &2.76$\times 10^{-2}$ &35.4 \\
SaN \cite{yu2015sketch} &512 &0.154 &0.225 & 0.53  &$7.98\times10^2$  &0.208 &0.292 & 0.21 &$2.85\times10^2$ \\
GN Triplet$^{*}$ \cite{sangkloy2016sketchy} &1024 &0.187 &0.301 &1.02  &$1.60\times10^3$   &0.529 &0.716 & 0.41 &$5.70\times10^2$  \\
3D shape$^{*}$ \cite{wang2015sketch} &64 &0.054 &0.072&7.53$\times 10^{-2}$  &99.8 $\mathrm{MB}$ &0.084  &0.079 & 2.64 $\times 10^{-2}$&35.6 \\
\hline
\tabincell{c}{Siamese-AlexNet} &4096 &0.367  &0.476 &5.35   &$6.39\times10^3$  &0.518 &0.690 & 1.68 &$2.28\times10^3$ \\
\hline
\tabincell{c}{Triplet-AlexNet}&4096 &0.448  &0.552 &5.35  &$6.39\times10^3$ &0.573 &0.761 & 1.68  $\mathrm{s}$&$2.28\times10^3$ \\
\hline
\hline
\multirow{3}{*}{\textbf{\tabincell{c}{DSH\\(Proposed)}}}  &32 (bits)&0.358 &0.486 &5.57$\times 10^{-4}$   &0.78 &0.653 &0.797 &2.55$\times 10^{-4}$  &0.28 \\
\cline{2-10}
&64 (bits)&0.521 &0.655 &7.03$\times 10^{-4}$  &1.56  &0.711 &0.858 &2.82$\times 10^{-4}$  &0.56 \\
\cline{2-10}
&128 (bits)&\textbf{0.570}& \textbf{0.694}&1.05$\times 10^{-3}$  &3.12  &\textbf{0.783} &\textbf{0.866} & 3.53$\times 10^{-4}$ &1.11 \\
\hline
\end{tabular}
}\scriptsize
\\'*' denotes we directly use the public models provided by the original papers without any fine-tuning on TU-Berlin Extension or Sketchy datasets.
\label{table:t11}
\vspace{-4ex}
\end{table*}

As shown in Fig.~\ref{learn}, we iteratively update $\mathbf{D}\rightarrow\mathbf{B}^{I}\rightarrow\mathbf{B}^{S}\rightarrow\{\Theta_{1}, \Theta_{2}\}$ in each epoch. As such,  DSH can be finally optimized within $T$ epochs in total, where $T=10\sim15$. Notice that the overall objective is lower-bounded, thus the convergence of  (\ref{e4}) is always guaranteed by coordinate descent used in our optimization. The overall DSH is summarized in Algorithm~\ref{aa1}.


Once the DSH model is trained, given a sketch query $\mathbf{S}_{q}$, we can compute its binary code $\mathbf{b}^{S_q}=\operatorname{sign}(\mathbf{F}_2(\mathbf{S}_{q};\Theta_{2}))$ with C2-Net (Top). For the retrieval database, the unified hash code of each image and sketch-token pair $\{\mathbf{I}, \mathbf{Z}\}$ is computed  as $\mathbf{b}^I=\operatorname{sign}(\mathbf{F}_1(\mathbf{I},\mathbf{Z};\Theta_{1},\Theta_{2}))$ with C1-Net (Bottom) and C2-Net (Middle).

\section{Experiments}
In this section, we conduct extensive evaluations of DSH on the two largest SBIR datasets: TU-Berlin Extension and Sketchy. Our method is implemented using Caffe\footnote{Our trained deep models can be downloaded from \texttt{https://github.com/ymcidence/DeepSketchHashing}.} with dual K80 GPUs for training our deep models and MATLAB 2015b on an i7 4790K CPU for binary coding.

\subsection{Datasets and Protocols}
\textbf{Datasets:}  \textbf{TU-Berlin} \cite{eitz2012humans} \textbf{Extension} contains 250 object categories with 80 free-hand sketches for each category. We also use 204,489 extended natural images  associated to TU-Berlin provided by \cite{zhang2016sketchnet} as our natural image retrieval gallery. \textbf{Sketchy} \cite{sangkloy2016sketchy} is a newly released dataset originally for fine-grained SBIR, in which 75,471 hand-drawn sketches of 12,500 objects (images) from 125 categories are included. To better fit the task of large-scale SBIR in our paper, we collect another 60,502 natural images (an average of 484 images/category) ourselves from ImageNet \cite{deng2009imagenet} to form a new retrieval gallery with  73,002 images in total. Similar to previous hashing evaluations, we randomly select 10 and 50 sketches from each category as the query sets for TU-Berlin  and Sketchy respectively, and the remaining sketches and gallery images\footnote{All natural images are used as both  training sets and  retrieval galleries.} are used for training.

\begin{table*}[ht]
\begin{center}
\newcommand{\tabincell}[2]{\begin{tabular}{@{}#1@{}}#2\end{tabular}}
\caption{Category-level SBIR using different cross-modality methods. For non-deep methods, 4096-d AlexNet \emph{fc7} image features  and 512-d SaN \emph{fc7} sketch features are used. For deep methods, raw natural images and sketches are used.}
\resizebox{0.999\textwidth}{!}{
\begin{tabular}{|c|c||c|c|c|c|c|c||c|c|c|c|c|c|}
  \hline
  \multicolumn{2}{|c||}{\multirow{3}{*}{\textbf{Method}}} &\multicolumn{6}{c||}{\textbf{TU-Berlin} \textbf{Extension}} & \multicolumn{6}{c|}{\textbf{Sketchy}}\\
\cline{3-14}
  & &\multicolumn{3}{c|}{\bf MAP} & \multicolumn{3}{c||}{\bf Precision@200} & \multicolumn{3}{c|}{\bf MAP} & \multicolumn{3}{c|}{\bf Precision@200} \\
  \cline{3-14}
  & &32 bits & 64 bits & 128 bits &  32 bits & 64 bits & 128 bits  & 32 bits & 64 bits & 128 bits &  32 bits & 64 bits & 128 bits \\
  \hline
  \hline
  \multirow{6}{*}{\tabincell{c}{Cross-Modality\\ Hashing Methods\\(binary codes)}}
  &CMFH \cite{ding2014collective} &0.149 &0.202 &0.180 &0.168 &0.282 &0.241 &0.320 &0.490 &0.190 &0.489 &0.657 &0.286\\
  &CMSSH  \cite{bronstein2010data}  &0.121 &0.183 &0.175 &0.143 &0.261 &0.233 &0.206 &0.211 &0.211 &0.371 &0.376 &0.375\\
  &SCM-Seq \cite{zhang2014large}  &0.211 &0.276 &0.332 &0.298 &0.372 &0.454 &0.306 &0.417&0.671 &0.442 &0.529 &0.758\\
  &SCM-Orth \cite{zhang2014large}&0.217 &0.301 &0.263 &0.312 &0.420 &0.470 &0.346 &0.536 &0.616 &0.467 &0.650 &0.776\\
  &CVH \cite{kumar2011learning}&0.214 &0.294 &0.318 &0.305 &0.411 &0.449 &0.325 &0.525 &0.624 &0.459 &0.641 &0.773\\
 &SePH  \cite{lin2015semantics}  &0.198 &0.270 &0.282 &0.307 &0.380 &0.398 &0.534 &0.607 &0.640 &0.694 &0.741 &0.768\\
  &DCMH \cite{jiang2016deep} &0.274 &0.382 &0.425 &0.332 &0.467 &0.540 &0.560 &0.622 &0.656 &0.730 &0.771 &0.784\\
   \hline
\textbf{Proposed}& \textbf{DSH}  &\textbf{0.358} &\textbf{0.521 }&\textbf{0.570} &\textbf{0.486} &\textbf{0.655} &\textbf{0.694} &\textbf{0.653} &\textbf{0.711} & \textbf{0.783}&\textbf{0.797} &\textbf{0.858} &\textbf{0.866}\\
  \hline\hline
  \multirow{4}{*}{\tabincell{c}{Cross-View Feature \\ Learning Methods\\(continuous-value vectors)}}
  &CCA \cite{thompson2005canonical} &0.276 &0.366 &0.365 &0.333 &0.482 &0.536 &0.361 &0.555 &0.705 &0.379 &0.610 &0.775\\
  &XQDA \cite{liao2015person} &0.191 &0.197 &0.201 &0.263 &0.278 &0.278 &0.460 &0.557 &0.550 &0.607 &0.715 &0.727\\
 \cline{3-14}
&PLSR \cite{wold1985partial} &\multicolumn{3}{c|}{0.141 (4096-d)} & \multicolumn{3}{c||}{0.215 (4096-d)} &\multicolumn{3}{c|}{0.462 (4096-d)} & \multicolumn{3}{c|}{0.623 (4096-d)} \\
  &CVFL \cite{xie2014cross} &\multicolumn{3}{c|}{0.289 (4096-d)} & \multicolumn{3}{c||}{0.407 (4096-d)} &\multicolumn{3}{c|}{0.675 (4096-d)} & \multicolumn{3}{c|}{0.803 (4096-d)} \\
  \hline
  \end{tabular}
  }
\label{table:t3}
\scriptsize
PLSR and CVFL are both based on reconstructing partial data to approximate full data, so the dimensions are fixed to 4096-d.
\vspace{-4ex}
\end{center}
\end{table*}
\begin{figure*}
  \centering
  \begin{tabular}{cccc}
     \includegraphics[width=0.229\textwidth]{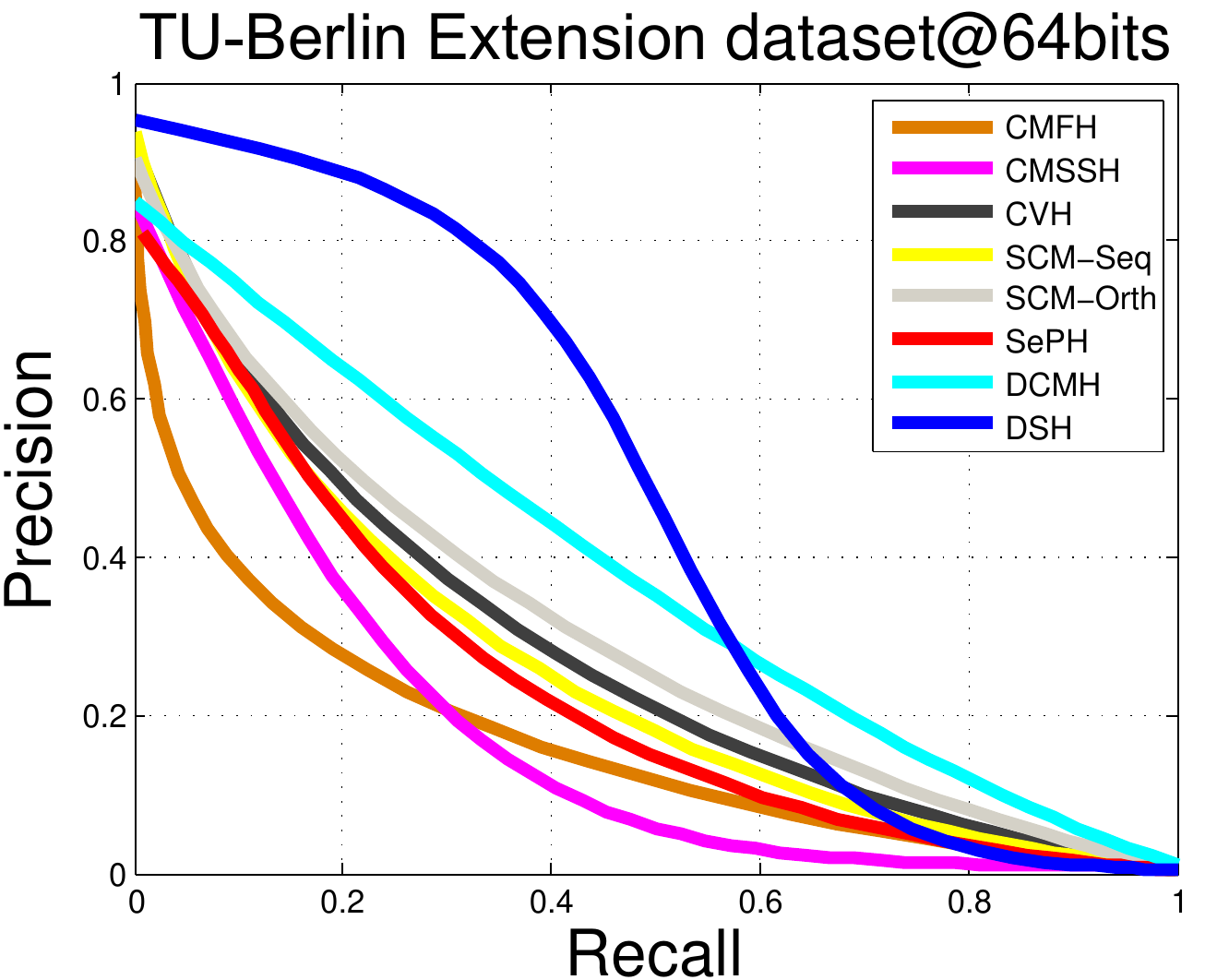} & \includegraphics[width=0.229\textwidth]{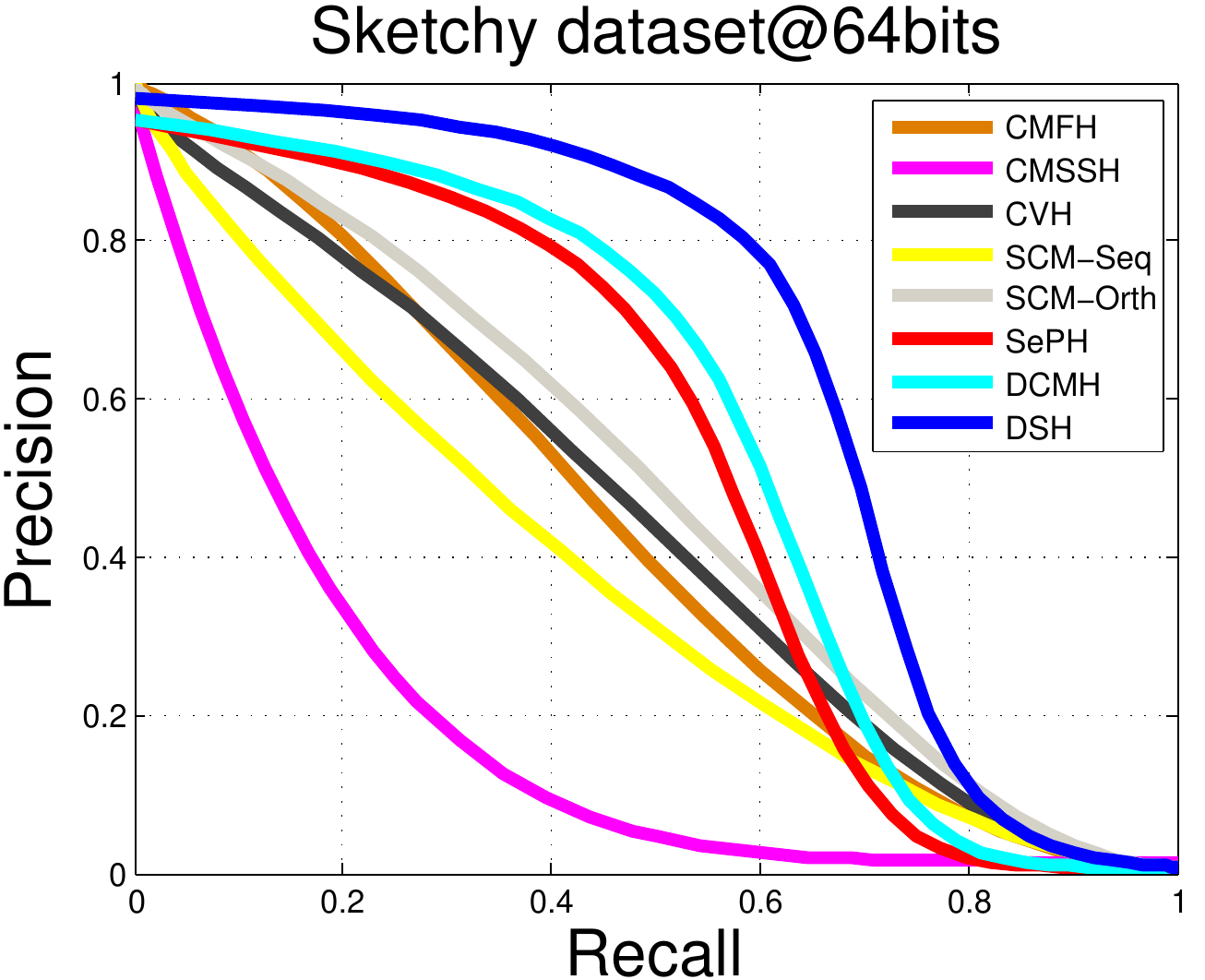} & \includegraphics[width=0.229\textwidth]{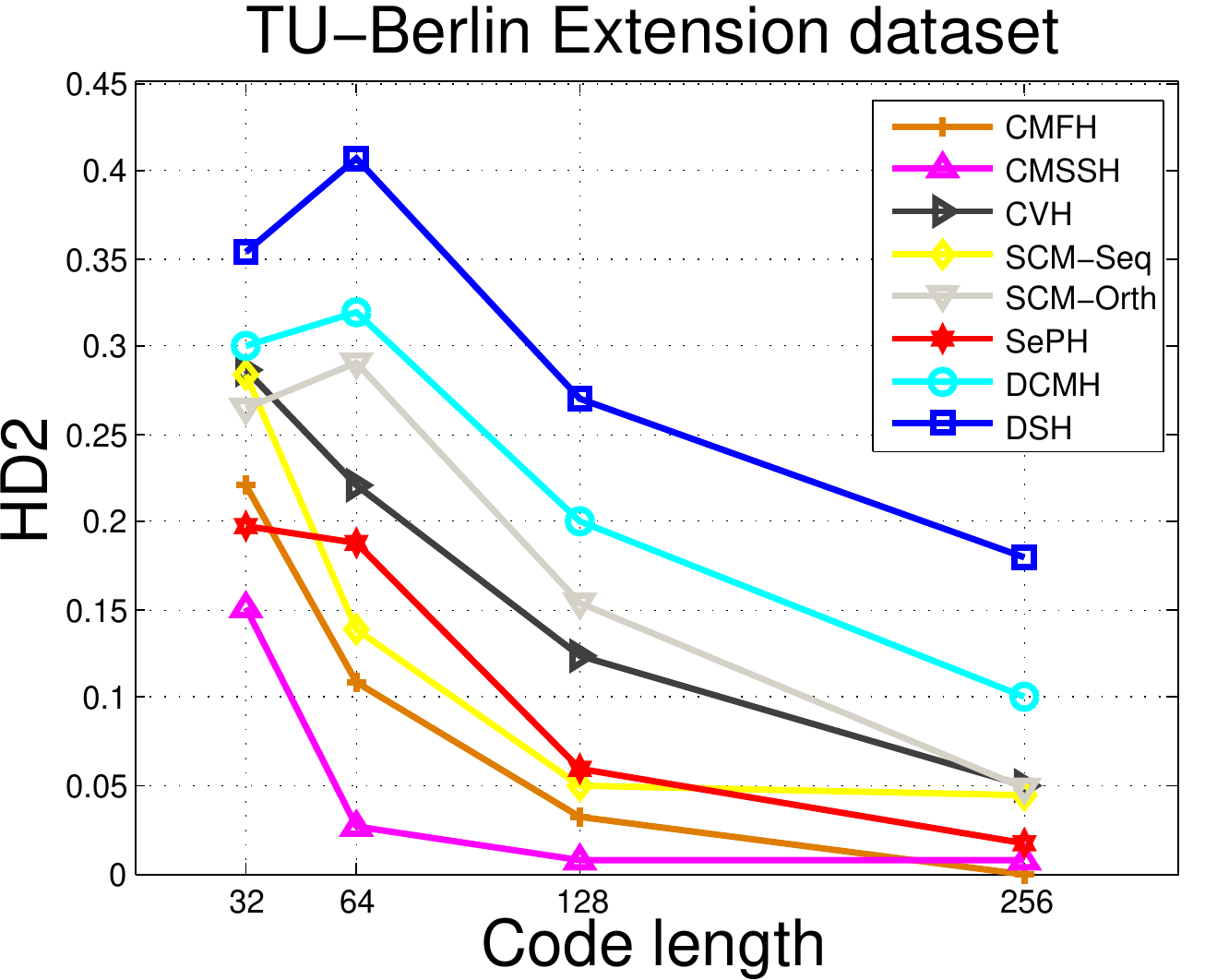} &
     \includegraphics[width=0.229\textwidth]{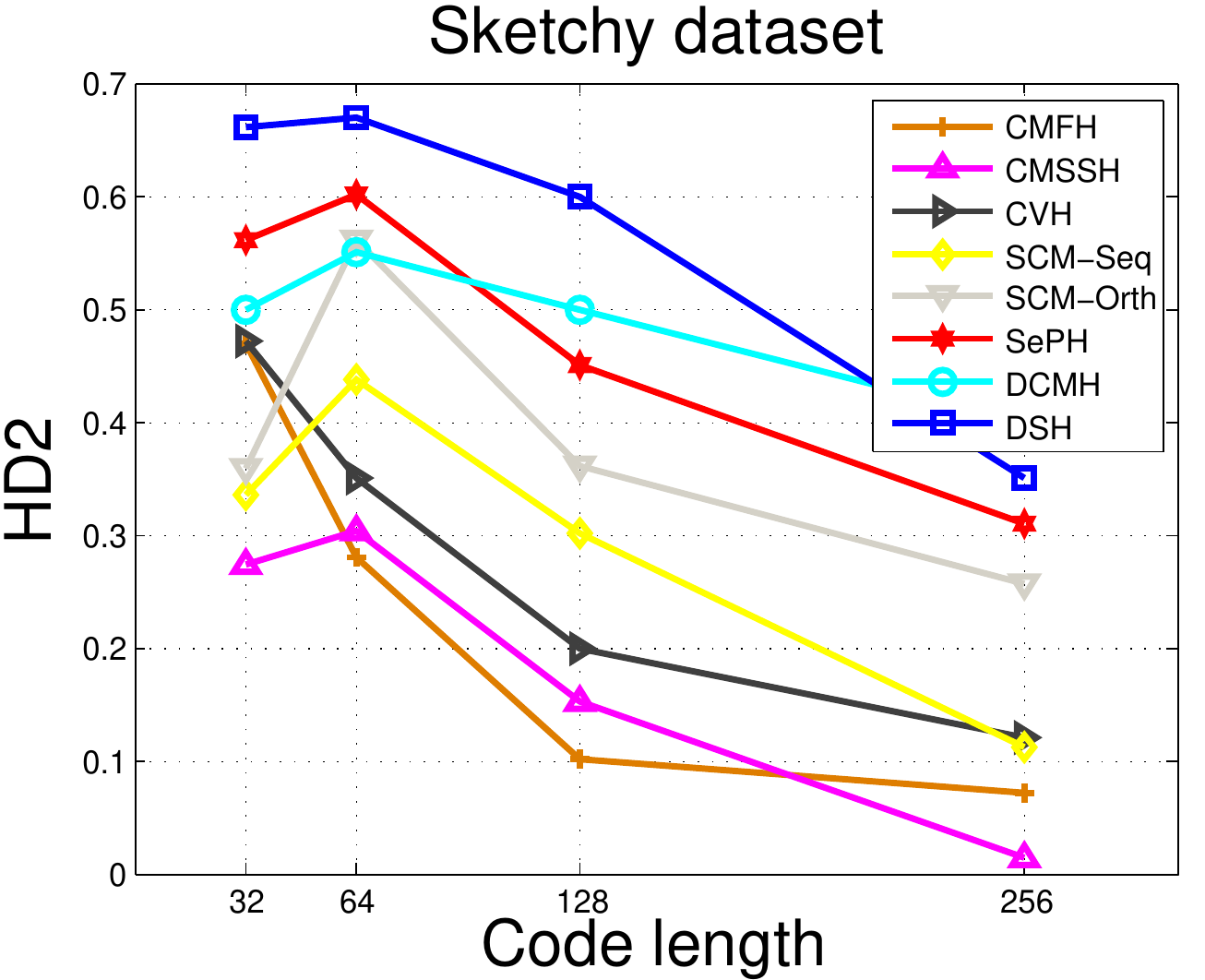}\\
  \end{tabular}
  \caption{Precision-recall curves and HD2 precision on both TU-Berlin Extension and Sketchy datasets.}
  \label{roc1}
\vspace{-3ex}
\end{figure*}

\vspace{1ex}
\textbf{Compared Methods and Implementation Details:} We first compare the proposed DSH with several previous SBIR methods, including hand-crafted HOG \cite{dalal2005histograms}, GF-HOG \cite{hu2010gradient},  SEHLO \cite{saavedra2014sketch}, LSK \cite{saavedrasketch}; and deep learning based Siamese CNN \cite{qi2016sketch}, Sketch-a-Net (SaN) \cite{yu2015sketch}, GN Triplet \cite{sangkloy2016sketchy}, 3D shape \cite{wang2015sketch}. For HOG, GF-HOG, SEHLO, Siamese CNN and 3D shape, we need first to compute Canny edgemaps from natural images and then extract the features. In detail,  we compute GF-HOG via a BoW scheme with a codebook size 3500; for HOG, SEHLO and LSK, we exactly follow the best settings used in \cite{saavedrasketch}.  Due to lack of stroke order information in the Sketchy dataset, we only use a single deep channel SaN in our experiments as in \cite{yusketch}. We fine-tune Siamese CNN and SaN on TU-Berlin and Sketchy datasets, while the public models of GN Triplet and 3D shape are only allowed for direct feature extraction without any re-training. Additionally, we add Siamese-AlexNet (with \emph{contrastive loss}) and Triplet-AlexNet (with \emph{triplet ranking loss}) as the baselines, both of which are constructed and trained by ourselves on two datasets. Particularly, the semantic pairwise/triplet supervision for our Siamese/Triplet-AlexNet are constructed  the same as \cite{qi2016sketch}/\cite{yaodeep} respectively.

\vspace{1ex}
Moreover, DSH is also compared with state-of-the-art cross-modality hashing techniques: Collective Matrix Factorization Hashing (CMFH) \cite{ding2014collective}, Cross-Modal Semi-Supervised Hashing (CMSSH) \cite{bronstein2010data}, Cross-View Hashing (CVH) \cite{kumar2011learning}, Semantic Correlation Maximization (SCM-Seq and SCM-Orth) \cite{zhang2014large}, Semantics-Preserving Hashing (SePH) \cite{lin2015semantics} and Deep Cross-Modality Hashing (DCMH) \cite{jiang2016deep}. Note that since DCMH is a deep hashing method originally for image-text retrieval, in our experiments, we modify it into a Siamese net by replacing the text embedding channel with an identical parallel image channel. In addition, another four cross-view feature embedding methods: CCA \cite{thompson2005canonical}, PLSR \cite{wold1985partial}, XQDA \cite{liao2015person}  and CVFL \cite{xie2014cross} are used for comparison. Except for DCMH, each image and sketch in both datasets are represented by 4096-d AlexNet \cite{krizhevsky2012imagenet} \emph{fc7} and 512-d SaN \emph{fc7} deep features, respectively. Since these hashing and feature embedding methods need pairwise data with corresponding labels as inputs, in our experiments, we further construct these deep features (extracted from TU-Berlin Extension/Sketchy datasets) into 100,000 sample pairs (with 800/400 pairs per category) to train all of the above cross-modality methods.

\vspace{1ex}
For the proposed DSH, we train our deep model using SGD on Caffe with an initial learning rate $\alpha$$=$0.001, momentum$=$0.9 and  batch size 64. We decrease $\alpha\rightarrow0.3\alpha$ every epoch and terminate the optimization after 15 epochs. For both datasets, our balance parameters are set to $\lambda$=$0.01$ and $\gamma$=$10^{-5}$ via cross validation on training sets.

In the test phase, we report the mean average precision (MAP) and precision at top-rank 200 (precision@200) to evaluate the category-level SBIR. For all hashing methods, we also evaluate the precision of Hamming distance with radius 2 (HD2) and the precision-recall curves. Additionally, we report the retrieval time per query ($\mathrm{s}$) from image galleries and memory loads (MB) for compared methods.

\begin{figure*}
  \centering
  \includegraphics[width=1\textwidth, height=0.26\textheight]{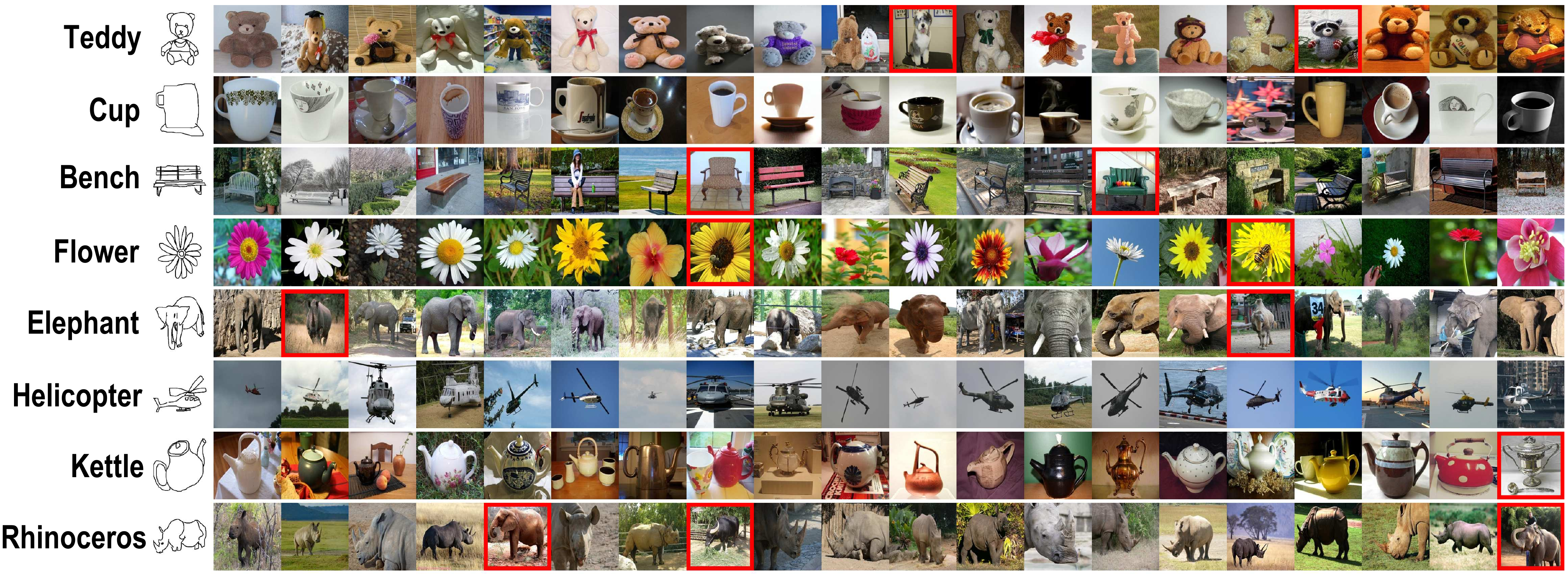}\\
  \caption{The visualization of our SBIR: eight example query sketches with their top-20 retrieval results on Sketchy dataset by using 128-bit DSH codes. Red boxes indicates false positives. (More examples can be seen in our \textbf{supplementary documents}.) }\label{sbir}
\vspace{-2.5ex}
\end{figure*}

\subsection{Results and Discussions}
\textbf{DSH vs. SBIR Baselines:}  In Table~\ref{table:t11}, we demonstrate the comparison of MAP and precision@200 over all SBIR methods on two datasets. Generally, deep learning-based methods can achieve much better performance than hand-crafted methods and the results on Sketchy are higher than those on TU-Berlin Extension since the data in Sketchy  is relatively simpler with fewer categories.  Our 128-bit DSH leads to  superior results with  0.138/0.142 and 0.210/0.105  improvements (MAP/precision@200) over the best-performing comparison methods on the two datasets, respectively. This is because the semi-heterogeneous deep architecture of DSH is specifically designed for category-level SBIR by  effectively introducing the auxiliary sketch-tokens to mitigate the geometric distortion between free-hand sketches and natural images. The other deep methods: Siamese CNN, GN Triplet and 3D shape only incorporate images and sketches as training data with a simple multi-channel deep structure. Among the compared methods, we notice 3D shape produces worse SBIR performance than  previous papers \cite{wang2015sketch,yusketch} reported. In \cite{yusketch}, the images from the retrieval gallery all contain \emph{well-aligned} objects with perfect \emph{background removal}, thus the edgemaps computed from such images can well represent the objects and have almost identical stroke patterns with free-hand sketches, which guarantees a good SBIR performance. However, in our tasks, all images in the retrieval gallery are \emph{realistic} with relatively \emph{complex backgrounds} and there is still a big dissimilarity between the computed edgemaps and sketches. Therefore, 3D shape features extracted from our edgemaps become ineffective. Similar problems also exist in SaN, HOG and  SHELO. In addition, the retrieval time and memory load are listed in Table~\ref{table:t11}. Our DSH can achieve significantly faster speed with much lower memory load compared to conventional SBIR methods during retrieval.

\begin{figure}
  \centering
  \begin{tabular}{cc}
     \includegraphics[width=0.225\textwidth]{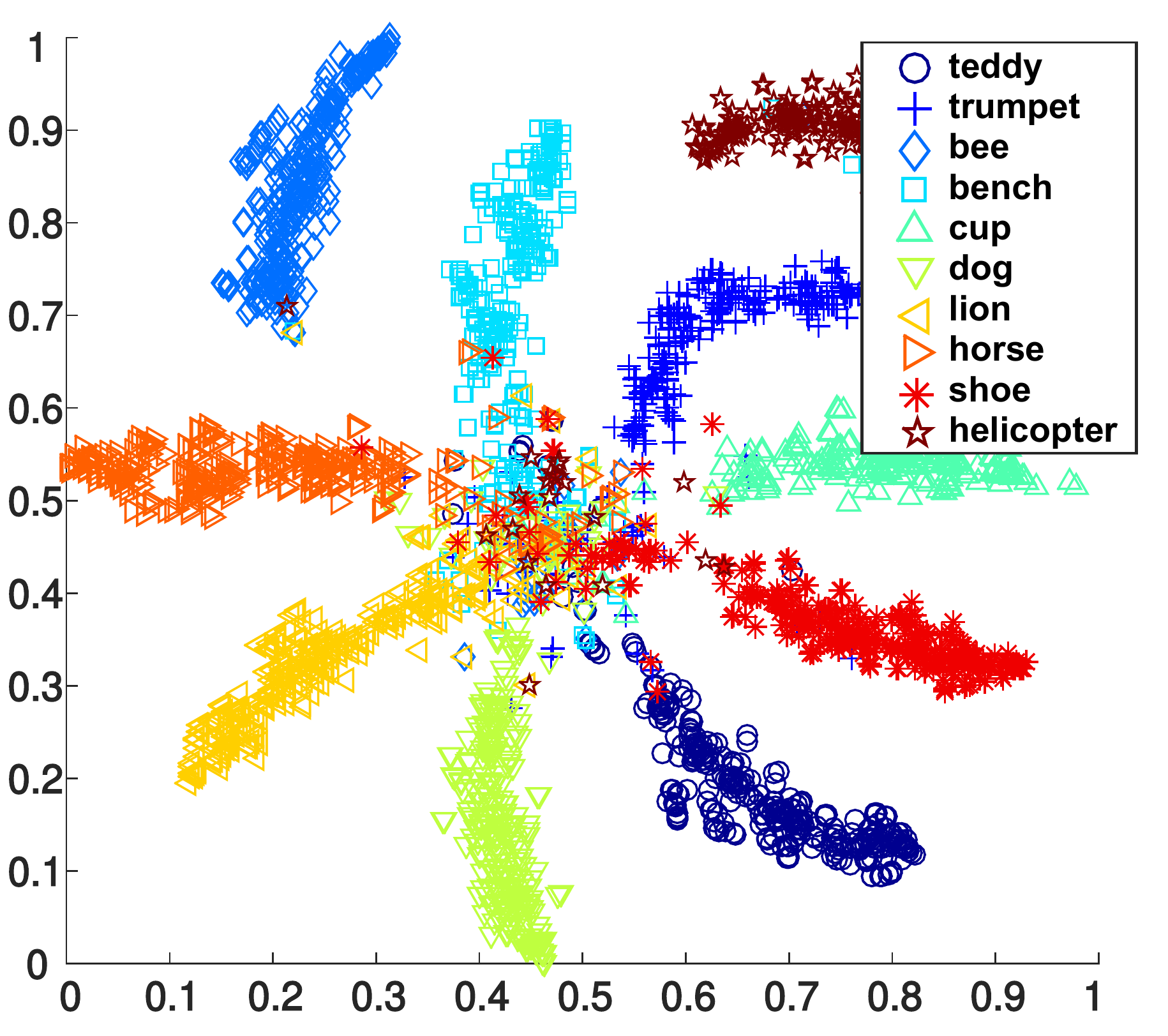} & \includegraphics[width=0.225\textwidth]{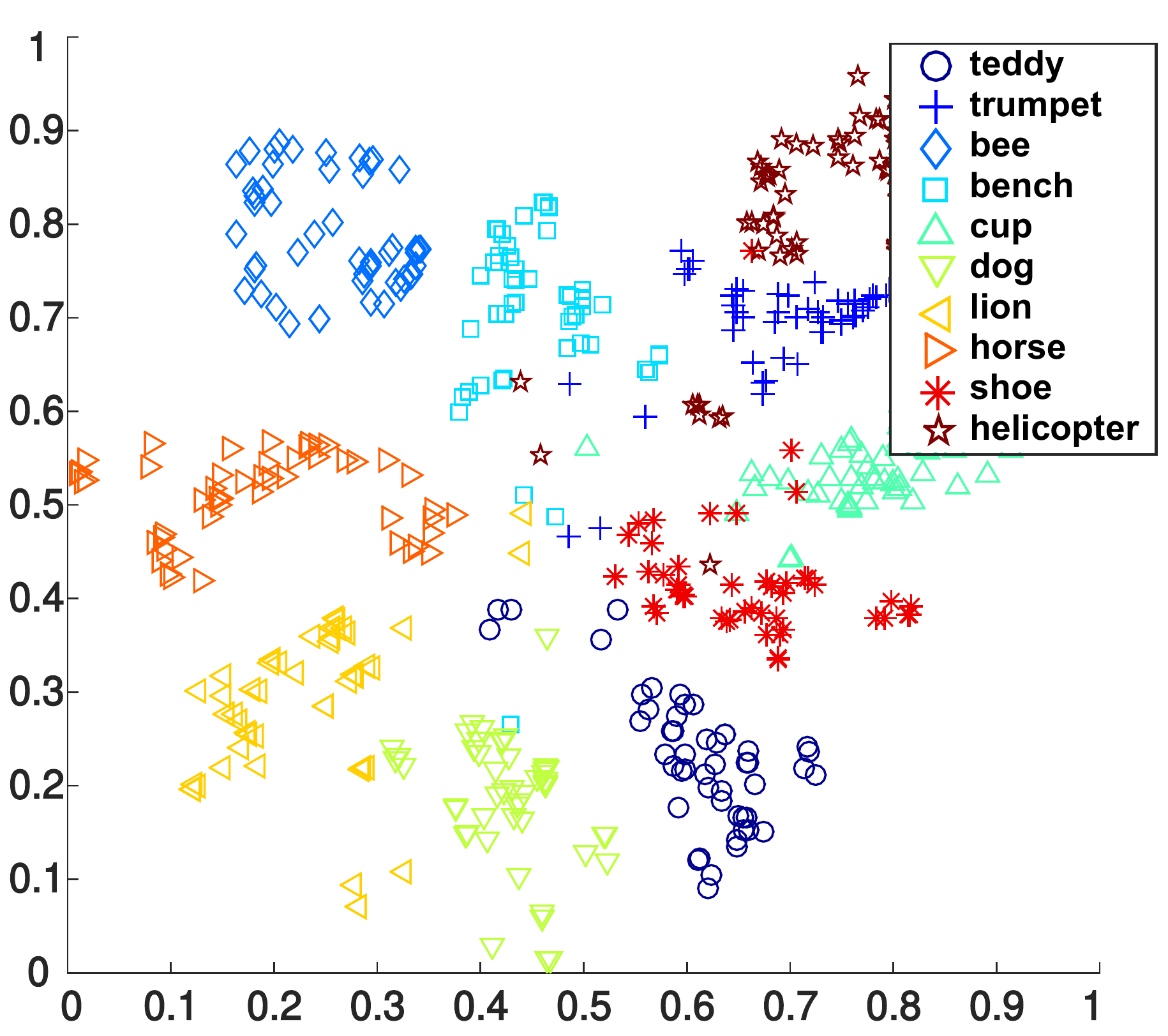} \\
     (a) \small Image retrieval gallery & (b) \small Test sketch queries \\
  \end{tabular}
  \vspace{0.1em}
  \caption{\emph{$t$-SNE} visualization of 32-bit DSH codes of  10 representative categories in the Sketchy dataset. After DSH,  natural images (a) and test sketch queries  (b) from the same categories are almost scattered into the same clusters. Meanwhile,  semantically similar categories  are distributed closely and otherwise far away.}  \label{tsne}
  \vspace{-0.1ex}
\end{figure}

\textbf{DSH vs. Cross-modality Hashing:} We also compare our DSH with cross-modality hashing/feature learning methods in Table~\ref{table:t3}. As mentioned before, we use the learned deep features as the inputs for  non-deep methods to achieve a fair comparison with our DSH. In particular, SCM-Orth  and SePH  always lead to high accuracies  among compared non-deep hashing methods on both datasets. With its deep end-to-end structure, DCMH can achieve better results than non-deep hashing methods, while CMFH and CMSSH produce the weakest results due to un(semi-)supervised learning mechanisms. For cross-view feature learning schemes, CCA and CVFL  achieve  superior performance on TU-Berlin Extension and Sketchy datasets,  respectively. Our DSH can consistently outperform all other methods in Table~\ref{table:t3}. The superior performance of DSH is also demonstrated in 64-bit precision-recall curves and HD2 curves along different code lengths (shown in Fig.~\ref{roc1}) by comparing the Area Under the Curve (AUC). Besides, we illustrate \emph{$t$-SNE} visualization in Fig.~\ref{tsne}  where the analogous DSH distributions of the test sketches and image gallery  intuitively reflect the effectiveness of DSH codes. Lastly, some query examples with top-20 SBIR retrieval results are shown in Fig.~\ref{sbir}.\vspace{2ex}

\textbf{DSH Component Analysis:} We have evaluated the effectiveness of different components of DSH in Table~\ref{table:t2}. Specifically, we construct a heterogeneous deep net by only using C2-Net (Top) and C1-Net (Bottom) channels with the same binary coding scheme. It produces around $0.073$ and $0.101$ MAP decreases by only using images and sketches on the respective datasets, which sufficiently proves the importance of sketch-tokens in order to mitigate the geometric distortion. We also observe that only using either the cross-view pairwise loss term or the semantic factorization loss term will result in worse performance than applying the full model, since the cross-view similarities and  the intrinsic semantic correlations captured in DSH can complement each other and simultaneously benefit the final MAPs. 
\begin{table}[t]
\newcommand{\tabincell}[2]{\begin{tabular}{@{}#1@{}}#2\end{tabular}}
\caption{Effectiveness (MAP 128 bits) of different components.}
\vspace{-4ex}
\begin{center}
\resizebox{0.5\textwidth}{!}{
\begin{tabular}{c|c|c}
\hline
\textbf{Method}&\tabincell{c}{\textbf{TU-Berlin}\\ \textbf{Extension}} & \textbf{Sketchy}\\
\hline
\hline
C2-Net (Top) + C1-Net (Bottom) \emph{only} &0.497 &0.682 \\
C2-Net (Top) + C2-Net (Middle)  \emph{only} &0.379 &0.507 \\
\hline
Using Cross-view Pairwise Loss \emph{only} &0.522 &0.715 \\
Using Semantic Factorization Loss \emph{only} &0.485 &0.667 \\
\hline
\textbf{Our proposed full DSH model}&\textbf{0.570} &\textbf{0.783} \\
\hline
\end{tabular}
}
\end{center}
\label{table:t2}
\vspace{-2.5ex}
\end{table}

\vspace{-1ex}
\section{Conclusion}
In this paper, we proposed a novel deep hashing framework, named deep sketch hashing (DSH), for fast sketch-based image retrieval (SBIR). Particularly, a semi-heterogeneous deep architecture was designed to encode free-hand sketches and natural images, together with the auxiliary sketch-tokens which can effectively mitigate the geometric distortion between the  two modalities. To train DSH,  binary codes and deep hash functions were jointly optimized in an alternating manner. Extensive experiments validated the superiority of DSH over the state-of-the-art methods in terms of  retrieval accuracy and time/storage complexity.

{\small
\bibliographystyle{ieee}
\bibliography{dsh_ref}
}

\end{document}